\documentclass{bmvc2k}

\usepackage{wrapfig}
\usepackage{graphicx}
\usepackage{amsmath}
\usepackage{amssymb}

\usepackage{multirow}
\usepackage{booktabs}
\usepackage{tabularx}
\newcommand{\bg}[1]{\boldsymbol{#1}} %Bold Greek letters
\newcommand{\bm}[1]{\mathbf{#1}} %Bold vectors and matrices

\newcommand\T{{\mathpalette\raiseT\intercal}}
\newcommand\raiseT[2]{%
\setbox0\hbox{$#1{#2}$}\raise\dp0\box0}

\title{LuKAN: A Kolmogorov-Arnold Network Framework for 3D Human Motion Prediction}

\addauthor{Md Zahidul Hasan}{zadidhasan11@gmail.com}{1}
\addauthor{A. Ben Hamza}{hamza@ciise.concordia.ca}{1}
\addauthor{Nizar Bouguila}{nizar.bouguila@concordia.ca}{2}

\addinstitution{
 Concordia University \\
 Montreal, QC, Canada
}

\runninghead{Hasan \textit{et al.}}{LuKAN for 3D Human Motion Prediction}

\begin{document}

\maketitle

\begin{abstract}
The goal of 3D human motion prediction is to forecast future 3D poses of the human body based on historical motion data. Existing methods often face limitations in achieving a balance between prediction accuracy and computational efficiency. In this paper, we present LuKAN, an effective model based on Kolmogorov-Arnold Networks (KANs) with Lucas polynomial activations. Our model first applies the discrete wavelet transform to encode temporal information in the input motion sequence. Then, a spatial projection layer is used to capture inter-joint dependencies, ensuring structural consistency of the human body. At the core of LuKAN is the Temporal Dependency Learner, which employs a KAN layer parameterized by Lucas polynomials for efficient function approximation. These polynomials provide computational efficiency and an enhanced capability to handle oscillatory behaviors. Finally, the inverse discrete wavelet transform reconstructs motion sequences in the time domain, generating temporally coherent predictions. Extensive experiments on three benchmark datasets demonstrate the competitive performance of our model compared to strong baselines, as evidenced by both quantitative and qualitative evaluations. Moreover, its compact architecture coupled with the linear recurrence of Lucas polynomials, ensures computational efficiency. Code is available at: \textcolor{blue}{https://github.com/zadidhasan/LuKAN}
\end{abstract}

\section{Introduction}
The task of 3D human motion prediction is to forecast the future 3D poses of a human body over a specified time horizon based on historical motion data. It empowers diverse applications requiring dynamic and responsive interaction with human movements, including human-object interaction~\cite{Diller2024CGHOI}, animation~\cite{Azadi2023Animation}, and autonomous driving~\cite{Djuric2020Driving,Wu2020MotionNet}. In recent years, substantial progress has been made in 3D human motion prediction~\cite{mao2019learning,dang2021msrgcn,ma2022progressively,Arij2022MotionMixer,Maosen2022SPGSN,guo2023mlp,Sun2023DeFeeNet,Wei2024NeRMo}, yet accurately forecasting future motion remains a major challenge due to the intrinsic complexity and variability of human movements. The spatio-temporal nature of human motion further compounds these challenges, requiring models to effectively capture both spatial inter-joint relationships and temporal dynamics across sequential frames.

State-of-the-art methods have embraced diverse neural network architectures tailored to the spatio-temporal nature of motion data, including Recurrent Neural Networks (RNNs)~\cite{Fragkiadaki2015RecurrentNM,jain2016structural,martinez2017human}, Graph Convolutional Networks (GCNs)~\cite{mao2019learning,mao2020history,dang2021msrgcn,zhang2023generating,ma2022progressively}, Transformers~\cite{mao2020history,Cai2020learnprogressive,Aksan2021spatiotemporal}, and Multi-Layer Perceptrons (MLPs)~\cite{Arij2022MotionMixer,guo2023mlp}. RNNs excel at modeling sequential dependencies but struggle with long-term sequences. GCN-based approaches capture spatial relationships through graph convolutions, but are prone to oversmoothing. Transformers, leveraging the self-attention mechanism, have quadratic computational complexity with respect to sequence length, requiring substantial computation for effective training. MLP-based models achieve reduced computational overhead, but use fixed activation functions and require deep architectures to model complex relationships. More recently, Kolmogorov-Arnold networks (KANs) have emerged as a compelling alternative to MLPs, demonstrating superior performance in function representation across various tasks, including regression~\cite{Liu2024KAN}, while mitigating spectral bias~\cite{Wang2024KANs}. Unlike MLPs, KANs leverage learnable activation functions on the edges. Existing GCN- and MLP-based approaches employ the discrete cosine transform (DCT) to encode motion in the frequency domain~\cite{mao2019learning,guo2023mlp}. However, the reliance on DCT may limit their flexibility in capturing localized motion patterns. Moreover, most GCN- and Transformer-based models incorporate MLPs as their core components for feature learning, inheriting a fundamental drawback of MLPs, namely spectral bias~\cite{Rahaman2019Bias}.

\smallskip\noindent\textbf{Proposed Work and Contributions}. In this paper, we propose LuKAN, a robust model for 3D human motion prediction based on KANs. It integrates a KAN layer that learns univariate functions parameterized by Lucas polynomials to capture interactions between temporal patterns across joints, and spatial projections that model inter-joint relationships. We summarize our contributions as follows: (1) We propose a novel architecture, leveraging KANs and the discrete wavelet transform to encode temporal information in the motion sequence by decomposing the trajectory of each body joint into low-frequency (coarse-scale) components and high-frequency components (fine-scale). Wavelet functions excel at capturing transient and rapidly changing features in a signal, offering a significant advantage over DCT, particularly for motion data where localized variations and dynamic changes are crucial. For instance, rapid hand gestures (high-frequency components) can be captured at fine scales, while slower, more gradual movements like walking (low-frequency components) can be captured at coarser scales; (2) We design a Temporal Dependency Learner to model both localized motion variations and global trends in human motion; (3) We conduct extensive experiments on benchmark datasets, showing that LuKAN achieves competitive performance with minimal computational overhead.

\section{Related Work}
\noindent\textbf{RNN-based Methods}.\quad RNNs have been extensively used in the early stages of human motion prediction research due to their ability to model temporal dependencies in sequential data~\cite{Fragkiadaki2015RecurrentNM,jain2016structural,martinez2017human,li2018convolutional}. These models excel at capturing temporal patterns, making them suitable for tasks where the sequence order and history play a vital role. However, RNN-based methods are often limited by their inability to effectively capture long-term dependencies and are prone to gradient instability during training, particularly for complex motion sequences.

\smallskip\noindent\textbf{GCN- and MLP-based Methods}.\quad GCNs represent human poses as graphs, with joints as nodes and bones as edges. This graph structure enables GCN-based methods to encode inter-joint dependencies naturally. Mao \emph{et al.}~\cite{mao2019learning} proposed a spatio-temporal network that applies DCT to input motion sequences to encode the temporal dynamics of joint coordinates in the trajectory space. The network uses GCNs with learnable adjacency matrices to capture spatial dependencies between body joints. Guo \emph{et al.}~\cite{guo2023mlp} introduced an effective approach using MLPs on the spatial and temporal dimensions of the DCT-transformed input. However, relying on DCT may constrain the flexibility of these models in capturing localized motion patterns effectively. Moreover, MLPs use fixed activation functions at their nodes, limiting their flexibility to adapt to diverse data patterns. Feng \textit{et al.}~\cite{Fenga2024MotionWavelet} introduced MotionWavelet, leveraging 2D wavelet transforms to model human motion patterns in the spatial-frequency domain. However, its reliance on diffusion models with guidance mechanisms to control prediction refinement results in higher computational cost. Our proposed LuKAN framework differs from existing methods in that it employs learnable 1D functions on its edges, allowing the network to adaptively model complex temporal dependencies in motion data. It also employs DWT to encode temporal dependencies in the joint trajectory, allowing the model to capture both coarse and fine-grained motion patterns. While both our model and MotionWavelet leverage wavelet transforms for human motion prediction, they differ significantly in terms of their architectural design and learning methodology. Unlike MotionWavelet~\cite{Fenga2024MotionWavelet}, which modifies motion signals repeatedly through the diffusion process, LuKAN retains high-frequency details. Moreover, using a KAN layer parameterized with Lucas polynomials provides flexibility and computational efficiency, as they are more efficient to evaluate than the piecewise construction of B-splines used in standard KANs.

\section{Method}
In this section, we first describe the task at hand. Next, we provide a preliminary background on KANs~\cite{Liu2024KAN,Wang2024KANs}. Then, we introduce the key building blocks of our network architecture.

\smallskip\noindent\textbf{Problem Description.}\quad Let $\bm{X}_{1:L}=(\bm{x}_{1},\dots,\bm{x}_{L})^{\T}\in\mathbb{R}^{L\times K}$ be a history motion sequence of $L$ consecutive 3D human poses, where $L$ is the look-back window, $K=3J$ in the feature dimension, and $J$ is the total number of body joints. At each time step $t$, each  pose $\bm{x}_{t}\in\mathbb{R}^{1\times K}$  is a flattened vector formed by concatenating the 3D coordinates of all joints in a single frame. The objective is to construct a predictive model that estimates a motion sequence $\hat{\bm{X}}_{L+1:L+T}=(\hat{\bm{x}}_{L+1},\dots,\hat{\bm{x}}_{L+T})\in\mathbb{R}^{T\times K}$ for the subsequent $T$ timesteps. To this end, we design an efficient model based on Kolmogorov-Arnold networks~\cite{Liu2024KAN}.

\smallskip\noindent\textbf{Kolmogorov-Arnold Networks.}\quad KANs are inspired by the Kolmogorov-Arnold representation theorem~\cite{Braun2009KANTH,Johannes2021KANTH}, which states that any continuous multivariate function on a bounded domain can be represented as a finite composition of continuous univariate functions of the input variables and the binary operation of addition. A KAN layer is a fundamental building block of KANs~\cite{Liu2024KAN}, and is defined as a matrix of 1D functions $\bg{\Phi}=(\phi_{q,p})$, where each trainable activation function $\phi_{q,p}$ is defined as a weighted combination, with learnable weights, of a sigmoid linear unit (SiLU) function and a spline function. Given an input vector $\bm{x}$, the output of an $\mathsf{L}$-layer KAN is given by
\begin{equation}
\text{KAN}(\bm{x})=(\bg{\Phi}^{(\mathsf{L}-1)}\circ\dots\circ\bg{\Phi}^{(1)}\circ\bg{\Phi}^{(0)})\bm{x},
\end{equation}
where $\bg{\Phi}^{(\ell)}$ is a matrix of learnable functions associated with the $\ell$-th KAN layer.

\begin{figure*}[!htb]
\centering\includegraphics[width=4.5in]{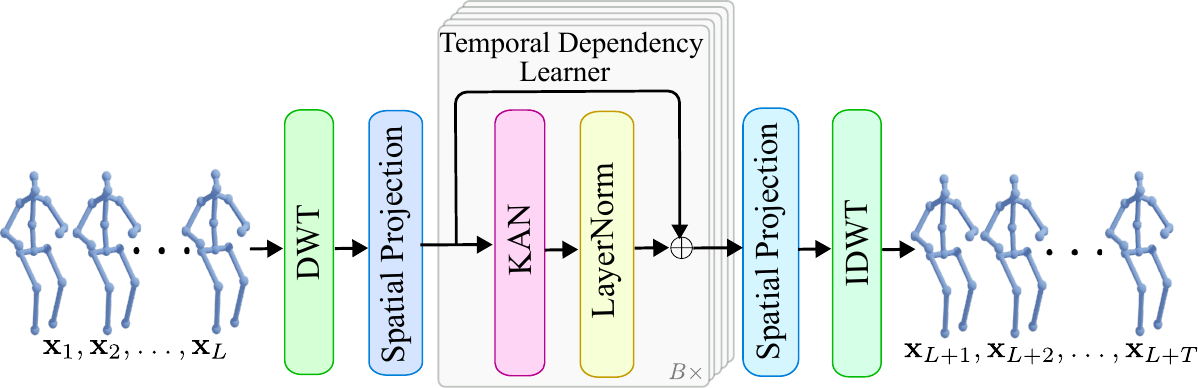}
\caption{\textbf{Overview of Model Architecture}. LuKAN processes input 3D motion data by applying DWT to encode temporal information. A spatial projection is applied both before and after the Temporal Dependency Learner block (repeated $B$ times). Each block consists of a KAN layer, LayerNorm, and a residual skip connection. The inverse DWT (IDWT) reconstructs the motion in the time domain, outputting a sequence of predicted 3D poses.}
\label{Fig:archirecture}
\end{figure*}

\smallskip\noindent\textbf{Model Architecture.}\quad The overall framework of our network architecture is depicted in Figure~\ref{Fig:archirecture}. LuKAN is designed to efficiently predict 3D human motion by modeling both spatial relationships and temporal dependencies in motion data.

\subsection{Temporal Encoding} \label{Subsec:IDWT}
\noindent\textbf{Joint Trajectory.}\quad  The $i$th column of the history motion sequence $\bm{X}_{1:L}$, denoted as $\bm{x}^{(i)}=(x_{1}^{(i)},\dots,x_{L}^{(i)})^{\T}$, represents the trajectory of the $i$-th skeleton joint over the $L$ consecutive frames in the sequence. The coordinates $x_{\ell}^{(i)}$ at each time step $\ell$ represent the position of the $i$-th joint in 3D space at that specific moment. This representation allows for capturing the motion of each joint individually over the observed time window.

\smallskip\noindent\textbf{Discrete Wavelet Transform Encoding.}\quad  To encode temporal information of the human motion in the trajectory, we employ DWT, which decomposes a signal into its approximate and detail components using wavelets, ensuring that localized temporal variations in the motion sequence are captured effectively. Specifically, given a wavelet (e.g., Daubechies wavelet), applying a three-level DWT to the $i$-th joint trajectory $\bm{x}^{(i)}$ yields
\begin{equation}
\bm{c}^{(i)}=\text{DWT}(\bm{x}^{(i)}),
\end{equation}
where $\bm{c}^{(i)}=(\bm{a}^{(i)},\bm{d}^{(i)})^{\T}$ is an $(L_{a}+L_{d})$-dimensional vector of wavelet coefficients that describe the signal's approximation and detail components. The approximation coefficients $\bm{a}^{(i)}\in\mathbb{R}^{L_{a}}$ represent the low-frequency (coarse-scale) components of the trajectory, while the detail coefficients $\bm{d}^{(i)}\in\mathbb{R}^{L_{d}}$ represent the high-frequency (fine-scale) variations in the trajectory. Unlike cosine waves, which oscillate indefinitely, wavelet functions are compact, with oscillations that diminish over time, enabling them to localize effectively and capture transient or rapidly changing features in a trajectory, which DCT cannot address as efficiently. The original trajectory can be reconstructed from its wavelet coefficients using the Inverse Discrete Wavelet Transform (IDWT) as follows:
\begin{equation}
\hat{\bm{x}}^{(i)}=\text{IDWT}(\bm{a}^{(i)},\bm{d}^{(i)}),
\end{equation}
which takes as input an $(L_{a}+L_{d})$-dimensional vector of wavelet coefficients and returns an $L$-dimensional reconstructed trajectory, ensuring that essential motion characteristics are preserved while enabling a more localized representation of human motion sequences.

\subsection{Spatial Projection}
The spatial projection maps the DWT-transformed history motion sequence into an embedding space of dimension $D$, capturing inter-joint dependencies and providing an expressive representation of the spatial structure of the human body. Its output is an $(L_{a}+L_{d})\times D$ matrix given by
\begin{equation}
\bm{Z}_1 = \text{DWT}(\bm{X}_{1:L})\bm{W}_1
\end{equation}
where $\bm{W}_{1}\in\mathbb{R}^{K\times D}$ is a learnable weight matrix, which defines a linear projection along the spatial (i.e., joint) dimension, and $D$ is the embedding dimension. For notational simplicity, the bias term is omitted here and throughout the following subsections.
%This spatial projection ensures that the spatial structure of the motion sequence is preserved before temporal dependencies are learned by the KAN layer.

\subsection{Temporal Dependency Learner}
The Temporal Dependency Learner is a core component of LuKAN, designed to capture temporal relationships within the motion sequence data. It operates as a sequence modeling block, emphasizing both local and global temporal dependencies to effectively predict future motion, while maintaining computational efficiency. This component consists of three key elements: a single KAN layer, LayerNorm, and a residual skip connection. The design choices are motivated as follows: (1) unlike MLPs, our proposed KAN effectively captures dependencies with its Lucas polynomials as learnable activation functions, providing flexibility in modeling both localized variations (such as fast changes in pose) and global trends (like slow transitions in motion), while reducing the need for excessively deep architectures; (2) LayerNorm helps stabilize training and ensures feature consistency across different motion sequences; and (3) a residual skip connection enhances gradient flow and prevents information loss, mitigating the limitations of purely feedforward architectures.

\medskip\noindent\textbf{KAN Layer.}\quad We employ a single KAN layer, with associated matrix $\bg{\Phi}=(\phi_{q,p})$ whose $(q,p)$-th entry is a function with learnable parameters. Each trainable function $\phi_{q,p}$ is parameterized by a weighted linear combination of Lucas polynomials
\begin{equation}
\bm{\phi}_{q, p}(x_{p}) = \sum_{r=0}^{R} \gamma_{q,p,r} P_{r}(x_{p}),
\end{equation}
where $x_p$ represents the $p$-th element of the joint trajectory vector, and $\gamma_{q,p,r}$ is the learnable coefficient of the $r$-th Lucas polynomial $P_{r}(x_p)$ for the $q$-th output element. These learnable parameters are adjusted during training to optimize the network's performance with the aim of improving the accuracy of the function approximation. In it important to mention that Lucas polynomials are defined recursively, making them computationally efficient to evaluate~\cite{Oruc2017Lucas}. Specifically, Lucas polynomials $P_{r}(x)$ are defined by the linear recurrence relation
\begin{equation}
P_{r}(x) = x P_{r-1}(x) + P_{r-2}(x),
\end{equation}
with initial conditions $P_{0}(x)=2$ and $P_{1}(x)=x$. The degree of $P_{r}(x)$ is equal to $r$.

\medskip\noindent\textbf{Layer Normalization (LN).}\quad LN is applied immediately after the KAN layer to standardize the output by normalizing feature activations.

\medskip\noindent\textbf{Residual Skip Connection.}\quad This skip connection links the input of KAN directly to its output, creating a residual pathway. Specifically, the output of the Temporal Dependency Learner is an $(L_{a}+L_{d})\times D$ matrix given by
\begin{equation}
\bm{Z_2} = \text{LN}(\text{KAN}(\bm{Z_1})) + \bm{Z_1},
\end{equation}
where KAN and LN are applied along the temporal dimension.

\subsection{Spatial Projection and Inverse Discrete Wavelet Transform}
The spatial projection, applied after the Temporal Dependency Learner, refines the spatial relationships between human body joints, ensuring structural consistency in the predicted poses. It models inter-joint dependencies, complementing the initial spatial projection. On the other hand, IDWT maps the temporally processed data back to the time domain. Together, the spatial projection and IDWT refine joint relationships and reconstruct the motion sequence in the time domain, resulting in an $L\times K$ output expressed as:
\begin{equation}
\bm{Z}_3 = \text{IDWT}(\bm{Z}_2\bm{W}_2),
\end{equation}
where $\bm{W}_2\in\mathbb{R}^{D\times K}$ is a learnable weight matrix. As pointed out in Subsection~\ref{Subsec:IDWT}, IDWT restores the temporal length from $L_a + L_d$ to the original $L$, thereby generating an $L\times K$ output $\bm{Z}_3$. The spatial projection corrects and reinforces joint relationships after KAN has processed the temporal dependencies, while IDWT ensures that these relationships are translated back into the time domain for motion reconstruction.

\medskip\noindent\textbf{Model Prediction.}\quad The predicted sequence is a $T\times K$ matrix given by
\begin{equation}
\hat{\bm{X}}_{L+1:L+T} = \widetilde{\bm{Z}}_{3} + \bm{X}_{L},
\end{equation}
where $T$ is the prediction horizon, $\widetilde{\bm{Z}}_{3}$ consists of the first $T$ rows of $\bm{Z}_3$, and $\bm{X}_{L}\in\mathbb{R}^{T\times K}$ is constructed by replicating the final pose $\bm{x}_L$ of the historical motion sequence $T$ times.

\medskip\noindent\textbf{Model Training.}\quad We train our model using the following loss function
\begin{equation}
\mathcal{L} = \frac{1}{T}\sum_{t = L+1}^{L+T} (\Vert\bm{x}_t - \hat{\bm{x}}_t\Vert_{2} + \Vert\bm{v}_t - \hat{\bm{v}}_t\Vert_{2}),
\end{equation}
where $\Vert\cdot\Vert$ denotes the $\ell_2$-norm, $\hat{\bm{x}}_{t}$ and $\bm{x}_{t}$ are the predicted and ground truth poses for the $t$-th predicted frame, $\bm{v}_t$ and $\hat{\bm{v}}_t$ are the associated velocities, respectively.

\section{Experiments}
\subsection{Experimental Setup}
\noindent\textbf{Datasets.}\quad We conduct experimental evaluations on three standard datasets: Human3.6M~\cite{Ionescu2014Human36M}, Archive of Motion Capture as Surface Shapes (AMASS)~\cite{Mahmood2019AMASS}, and 3D Pose in the Wild dataset (3DPW)~\cite{Timo20183DPW}. We follow standard protocols~\cite{mao2020history} for data preprocessing and splitting. Additional results and ablation studies are provided in the supplementary material.
%\begin{itemize}
%\item \textbf{Human3.6M} is among the most extensive and widely adopted datasets for 3D human motion analysis, comprising over 3.6 million frames of human activities recorded in a controlled indoor environment. It features 7 actors performing 15 everyday activities, including Walking, Sitting, Eating, Greeting, Discussing, and Taking Photos. The dataset is preprocessed to represent poses as 3D coordinates with 22 joints per frame. For evaluation, Subject 11 (S11) is reserved for validation, Subject 5 (S5) is used for testing, and the remaining 5 subjects are employed for training.
%
%\item \textbf{AMASS} is a comprehensive motion database, consolidating multiple optical marker-based motion capture datasets into a standardized framework via the Skinned Multi-Person Linear (SMPL) model. This standardization ensures uniform parameterization across diverse motion recordings. Comprising 51K frames of human activities in indoor and outdoor environments, each pose is represented by 23 joints. For evaluation, the AMASS-BMLrub dataset is used for testing.
%
%\item \textbf{3DPW} includes a diverse range of challenging human activities, spanning both controlled indoor settings and natural outdoor environments. To assess the robustness and generalization of our model trained on AMASS, we evaluate its performance on the 3DPW test set, where each pose is represented by 18 body joints.
%\end{itemize}

\smallskip\noindent\textbf{Evaluation Metric and Baselines.}\quad We assess the model's performance using the Mean Per Joint Position Error (MPJPE), measured in millimeters, where lower values correspond to better prediction performance. We benchmark LuKAN against several state-of-the-art approaches for 3D human motion prediction, including ConvSeq2Seq~\cite{li2018convolutional}, Learning Trajectory Dependencies (LTD)~\cite{mao2019learning}, History repeats (Hisrep)~\cite{mao2020history}, Dynamic Multiscale Graph Neural Networks (DMGNN)~\cite{Maosen2020DMGNN}, MultiScale Residual Graph Convolution Network (MSR-GCN)~\cite{dang2021msrgcn}, Spatial and Temporal Dense Graph Convolutional Network (ST-DGCN)~\cite{ma2022progressively}, Context-based Interpretable Spatio-Temporal Graph Convolutional Network (CIST-GCN)~\cite{Medina2024CISTGCN}, MotionMixer~\cite{Arij2022MotionMixer}, Skeleton-Parted Graph Scattering Networks (SPGSN)~\cite{Maosen2022SPGSN}, and Simple Multi-Layer Perceptron (SiMLPe)~\cite{guo2023mlp}.

\smallskip\noindent\textbf{Implementation Details.} \quad All experiments are performed on a single NVIDIA RTX 3070 GPU with 8GB of memory using PyTorch. Our model is trained for 50K epochs on Human3.6M and 115K epochs on AMASS, using a batch size of 128. We use Adam optimizer~\cite{Kingma2015Adam} with a weight decay of $10^{-4}$. The learning rate is initialized at $3\times 10^{-4}$ and decayed to $10^{-5}$ after 30K epochs. The look-back window is set to $L=50$, with a prediction horizon of $T=10$ for Human3.6M, and $T=25$ for AMASS and 3DPW. We employ Daubechies wavelets with 4 vanishing moments in both DWT and IDWT, and we set the number of levels of decomposition to 3. We also set the number of temporal dependency learner blocks to $B=48$.

\subsection{Results and Analysis}
\noindent\textbf{Results on Human3.6M.}\quad We report the MPJPE errors averaged across all time steps in Table~\ref{tab:all_baselines} for both short-term (80ms - 400ms) and long-term (560ms - 1000ms) predictions. The results demonstrate the effectiveness of LuKAN compared to the best-performing baseline, SiMLPe. LuKAN consistently achieves lower MPJPE errors across all time steps, with notable relative error reductions. For instance, at the 720ms prediction horizon, LuKAN achieves an MPJPE of 89.9mm compared to 90.1mm for SiMLPe, yielding a relative error reduction of approximately 0.22\%. Similarly, at the 1000ms horizon, LuKAN reduces the MPJPE to 109.3mm from SiMLPe's 109.4mm, resulting in a relative error reduction of approximately 0.09\%. These results highlight LuKAN's capability to improve upon the state-of-the-art, while maintaining its simple and efficient architecture.

\begin{table}[!htb]
\caption{Average MPJPE results of our model and baseline methods on Human3.6M for different prediction time steps in milliseconds (ms) ranging from 80ms to 1000ms. These MPJPE errors, measured in millimeters (mm), are averaged across all different actions in the dataset. The best results are shown in \textbf{bold}, and the second best results are \underline{underlined}.}
\label{tab:all_baselines}
\smallskip
\centering
\small
\setlength\tabcolsep{4pt} % Reduce column padding
\begin{tabular}{lcccccccc}
\toprule
\multirow{2}{*}{} & \multicolumn{8}{c}{MPJPE (mm)$\downarrow$} \\ \cline{2-9}
& 80 & 160 & 320 & 400 & 560 & 720 & 880 & 1000 \\
\midrule
% Repeating Last-Frame\cite{b15} & 23.8 & 44.4 & 76.1 & 88.2 & 107.4 & 121.6 & 131.6 & 136.6 \\
% One FC\cite{b15} & 14.0 & 33.2 & 68.0 & 81.5 & 101.7 & 115.1 & 124.8 & 130.0 \\
%Res. sup.\cite{ref:ressup} & 25.0 & 46.2 & 77.0 & 88.3 & 106.3 & 119.4 & 130.0 & 136.6 \\
ConvSeq2Seq~\cite{li2018convolutional} & 16.6 & 33.3 & 61.4 & 72.7 & 90.7 & 104.7 & 116.7 & 124.2 \\
% LTD-50-25\cite{b6} & 12.2 & 25.4 & 50.7 & 61.5 & 79.6 & 93.6 & 105.2 & 112.4 \\
LTD-10-10~\cite{mao2019learning} & 11.2 & 23.4 & 47.9 & 58.9 & 78.3 & 93.3 & 106.0 & 114.0 \\
Hisrep~\cite{mao2020history} & 10.4 & 22.6 & 47.1 & 58.3 & 77.3 & 91.8 & 104.1 & 112.1 \\
DMGNN~\cite{Maosen2020DMGNN} & 17.0 & 33.6 & 65.9 & 79.7 & 103 & - & - & 137.2 \\
MSR-GCN~\cite{dang2021msrgcn} & 11.3 & 24.3 & 50.8 & 61.9 & 80.0 &  - & -  & 112.9 \\
ST-DGCN~\cite{ma2022progressively} & 10.6 & 23.1 & 47.1 & 57.9 & \underline{76.3} & 90.7 & 102.4 & 109.7 \\
SPGSN~\cite{Maosen2022SPGSN} & 10.4 & 22.3 & 47 & 58.2 &  77.4 &  - &  - & 109.6\\
CIST-GCN~\cite{Medina2024CISTGCN} & 10.5 & 23.2 & 47.9 & 59.0 & 77.2 & - & - & 110.3\\
MotionMixer~\cite{Arij2022MotionMixer} & 11	& 23.6 &	47.8	& 59.3 &	77.8	& 91.4	& 106 &	111\\
SiMLPe~\cite{guo2023mlp} & \underline{9.6} & \underline{21.7} & \underline{46.3} & \underline{57.3} & \textbf{75.7} & \underline{90.1} & \underline{101.8} & \underline{109.4}\\
\midrule
% our model with DCT (Ours) & \textbf{9.4}	& \textbf{21.4} & \textbf{45.8}	& \textbf{56.8}	& \textbf{75.7}	& 90.2	& \underline{101.8}	& 109.5\\
LuKAN (ours) & \textbf{9.4}	& \textbf{21.5} & \textbf{46.2}	& \textbf{57.2}	& \textbf{75.7}	& \textbf{89.9}	& \textbf{101.6}	& \textbf{109.3}\\
\bottomrule
\end{tabular}
\end{table}

\smallskip\noindent\textbf{Results on AMASS and 3DPW.}\quad We train our model on the AMASS dataset and test it on on the AMASS-BMLrub and 3DPW datasets, adhering to the standard evaluation protocol outlined in~\cite{mao2020history}. The results in Table~\ref{tab:comparison} provide a comprehensive comparison of our model against strong baseline methods on the AMASS-BMLrub and 3DPW datasets, evaluated in terms of MPJPE across different prediction horizons. On AMASS-BMLrub, LuKAN achieves competitive results, particularly excelling in short-term predictions. At 80ms and 160ms, LuKAN matches the best-performing LTD-10-10 with MPJPEs of 10.6mm and 19.3mm, respectively. For longer horizons, LuKAN consistently demonstrates robust performance, achieving the second-best MPJPE scores, such as 34.4mm at 320ms and 66.4mm at 1000ms. Compared to SiMLPe at 320ms, for example, LuKAN yields comparable performance, highlighting its ability to stay on par with state-of-the-art models. On the more challenging 3DPW dataset, which evaluates the generalization ability of prediction models, LuKAN consistently outperforms all baselines across all prediction horizons. For instance, at 320ms, LuKAN achieves an MPJPE of 37.9mm, outperforming SiMLPe's 38.1mm with a relative error reduction of 0.52\%. At 1000ms, LuKAN achieves an MPJPE of 72.2mm, matching SiMLPe and further underscoring its robustness in generalization. Overall, the combination of competitive performance in short-term predictions and robust results in long-term horizons highlights LuKAN's versatility and ability to balance prediction accuracy and efficiency across different time horizons.

\begin{table*}[!htb]
\caption{\textbf{Performance comparison of our model and baselines on AMASS-BMLrub and 3DPW} for various prediction horizons.}
\label{tab:comparison}
\smallskip
\centering
\small
\setlength\tabcolsep{1.2pt} % Reduce column padding
\begin{tabular}{lcccccccc|cccccccc}
\toprule
\multirow{2}{*}{} & \multicolumn{8}{c}{AMASS-BMLrub} & \multicolumn{8}{c}{3DPW} \\
\cmidrule(lr){2-9} \cmidrule(lr){10-17}
& 80 & 160 & 320 & 400 & 560 & 720 & 880 & 1000 & 80 & 160 & 320 & 400 & 560 & 720 & 880 & 1000 \\
\hline
ConvSeq2Seq~\cite{li2018convolutional} & 20.6 & 36.9 & 59.7 & 67.6 & 79.0 & 87.0 & 91.5 & 93.5 & 18.8 & 32.9 & 52.0 & 58.8 & 69.4 & 77.0 & 83.6 & 87.8 \\
LTD-10-10~\cite{mao2019learning} & \textbf{10.3} & \textbf{19.3} & 36.6 & 44.6 & 61.5 & 75.9 & 86.2 & 91.2 & \underline{12.0} & \underline{22.0} & 38.9 & 46.2 & 59.1 & 69.1 & 76.5 & 81.1 \\
LTD-10-25~\cite{mao2019learning} & 11.0 & 20.7 & 37.8 & 45.3 & 57.2 & 65.7 & 71.3 & 75.2 & 12.6 & 23.2 & 39.7 & 46.6 & 57.9 & 65.8 & 71.5 & 75.5 \\
Hisrep~\cite{mao2020history} & 11.3 & 20.7 & 35.7 & 42.0 & 51.7 & 58.6 & 63.4 & 67.2 & 12.6 & 23.1 & 39.0 & 45.4 & \underline{56.0} & 63.6 & 69.7 & \underline{73.7} \\
SiMLPe~\cite{guo2023mlp} & 10.8 & \underline{19.6} & \textbf{34.3} & \textbf{40.5} & \textbf{50.5} & \textbf{57.3} & \textbf{62.4} & \textbf{65.7} & 12.1 & 22.1 & \underline{38.1} & \underline{44.5} & \textbf{54.9} & \underline{62.4} & \underline{68.2} & \textbf{72.2} \\\hline
Ours & \underline{10.6} & \textbf{19.3}	& \underline{34.4}	& \underline{40.8} & \underline{50.9} & \underline{57.6} & \underline{62.7} & \underline{66.4} & \textbf{11.9} & \textbf{21.8} & \textbf{37.9} & \textbf{44.4} & \textbf{54.9}	& \textbf{62.2}	& \textbf{68.1}	& \textbf{72.2}\\
\bottomrule
\end{tabular}
\end{table*}

%\smallskip\noindent\noindent\textbf{Model Efficiency.}\quad At 1000ms, for instance, our model uses only 0.045M parameters, substantially fewer than SiMLPe's 0.14M, and requires just 2.1M FLOPs, compared to SiMLPe's 8.76M, all while achieving better prediction results.

\begin{figure}[!htb]
\centering\includegraphics[width=3.52in]{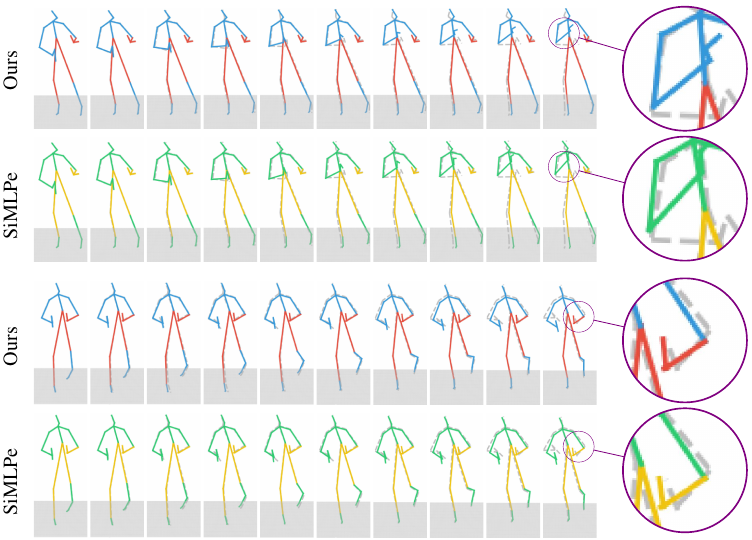}
\caption{\textbf{Visual comparison results of our model and the SiMLPe baseline} on two actions: Directions (top) and Eating (bottom). Predicted poses from our model are depicted in red and blue, while those from SiMLPe are shown in yellow and green. Ground truth poses, represented by dashed lines, are overlaid with the predictions to highlight deviations.}
\label{fig:viz}
\end{figure}

\smallskip\noindent\noindent\textbf{Qualitative Results.}\quad In Figure \ref{fig:viz}, we present a comparison of our predicted poses with those generated by SiMLPe for the Directions and Eating actions from Human3.6M. To facilitate visual assessment, the predicted frames are overlaid on the ground truth poses, highlighting any deviations. For both actions, our model demonstrates superior alignment with the ground truth, particularly for the Directions action. Notably, the predicted leg positions from our model are closer to the ground truth compared to those predicted by SiMLPe.

\subsection{Ablation Study}
\noindent\textbf{Effect of Temporal Encoding.}\quad The results in Table~\ref{tab:comparison_methods} compare the performance of DWT and DCT for temporal encoding across Human3.6M, AMASS, and 3DPW datasets in terms of MPJPE. On Human3.6Mt, DWT achieves an MPJPE of 89.9mm at 720ms, outperforming DCT's 90.2mm with a relative error reduction of 0.33\%. Similarly, at 1000ms, DWT achieves a lower MPJPE of 109.3mm compared to DCT's 109.5mm, yielding a relative error reduction of 0.18\%. On AMASS, DWT consistently outperforms DCT across all time steps. For instance, at 400ms, DWT achieves an MPJPE of 40.8mm compared to 41.5mm for DCT, resulting in a relative error reduction of 1.69\%. At 1000ms, DWT reduces the MPJPE to 66.4mm compared to DCT's 66.9mm, with a relative error reduction of 0.75\%. On 3DPW, the difference between DWT and DCT is less pronounced, but DWT achieves slightly better results for most time steps. At 320ms, DWT achieves an MPJPE of 37.9mm compared to 38.4mm for DCT, yielding a relative error reduction of 1.3\%. Overall, DWT demonstrates consistent improvements over DCT across all datasets, particularly in short-term predictions.
% highlighting its ability to capture localized temporal information more effectively.

%\begin{wraptable}{R}{5.5cm}
\begin{table}[!htb]
\caption{Ablation study on the choice of temporal encoding: DWT vs. DCT across all datasets for various prediction horizons. DWT consistently outperforms DCT.}
\label{tab:comparison_methods}
%\smallskip
\centering
\small
\setlength\tabcolsep{3.9pt} % Reduce column padding
\begin{tabular}{llcccccccc}
\toprule
\multirow{2}{*}{} & \multirow{2}{*}{} & \multicolumn{8}{c}{MPJPE (mm)$\downarrow$} \\
\cmidrule(lr){3-10}
& & 80 & 160 & 320 & 400 & 560 & 720 & 880 & 1000 \\
\midrule
\multirow{2}{*}{Human3.6M} & DCT & \textbf{9.4}	& \textbf{21.4}	& \textbf{45.8}	& \textbf{56.8}	& \textbf{75.7}	& 90.2	& 101.8	& 109.5\\
& DWT &\textbf{9.4} & 21.5 & 46.2 & 57.2 & \textbf{75.7} & \textbf{89.9} & \textbf{101.6} & \textbf{109.3}\\
\midrule
\multirow{2}{*}{AMASS} & DCT & 10.9 & 19.7 & 34.9 & 41.5 & 51.6 & 58.7 & 63.6 & 66.9 \\
& DWT & \textbf{10.6} &	\textbf{19.3}	& \textbf{34.4}	& \textbf{40.8}	& \textbf{50.9}	& \textbf{57.6} &	\textbf{62.7}	& \textbf{66.4} \\
\midrule
\multirow{2}{*}{3DPW} & DCT & 12.2 & 22.2 & 38.4 & 44.9 & 55.1 & 62.3 & \textbf{68.1} & \textbf{72.2} \\
& DWT & \textbf{11.9}	& \textbf{21.8} &	\textbf{37.9}	& \textbf{44.4}	& \textbf{54.9}	& \textbf{62.2} &	\textbf{68.1}	& \textbf{72.2}\\
\bottomrule
\end{tabular}
\end{table}
%\end{wraptable}

\medskip\noindent\textbf{Effect of Polynomial Basis.} \quad The results in Table~\ref{tab:kan_comparison} highlight the superior performance of Lucas polynomials compared to B-splines, used in standard KANs, and other polynomial bases. At 400ms, Lucas polynomials outperform the next best basis, Hermite, yielding a relative error reduction of 0.17\%. Similarly, at 1000ms, Lucas polynomials achieve an MPJPE of 109.3mm, outperforming Hermite's 110.1mm by a relative reduction of 0.73\%. In comparison to B-splines, the improvements are more pronounced, yielding a relative error reduction of 2.5\%. At 320ms, Lucas polynomials achieve an MPJPE of 46.2mm compared to 49.0mm for B-splines, resulting in a relative error reduction of 5.71\%. These results demonstrate that Lucas polynomials yield significant improvements over B-splines and other polynomial bases, for both short- and long-term predictions.
\begin{table}[!htb]
\caption{Ablation study on the choice of the polynomial basis in KAN for various prediction horizons. Lucas polynomials yield significant improvements over B-splines.}
\label{tab:kan_comparison}
\smallskip
\centering
\small
\setlength\tabcolsep{3.2pt} % Reduce column padding
\begin{tabular}{lcccccccc}
\toprule
\multirow{2}{*}{Polynomials} & \multicolumn{8}{c}{MPJPE (mm)$\downarrow$} \\
\cmidrule(lr){2-9}
& 80 & 160 & 320 & 400 & 560 & 720 & 880 & 1000 \\
\midrule
B-Splines & 10.3 & 23.3 & 49.0 & 60.1 & 78.7 & 92.7 & 104.5 & 112.1\\
%B-Splines (RBF) & 10.5& 23.5 & 49.1 & 60.3 & 78.6 & 92.5 & 104.1 & 111.7 \\
Chebyshev  & 9.7 & 22.1 & 47.2 & 58.3 & 77.1 & 91.7 & 103.7 & 111.6 \\
Legendre  & 9.6 & 21.8 & 46.8 & 57.9 & 76.3 & 90.4 & 102.4 & 110.1 \\
Hermite  & 9.5 & 21.6 & 46.3 & 57.3 & 76.0 & 90.3 & 102.2 & 110.1\\
Lucas  & \textbf{9.4} & \textbf{21.5} & \textbf{46.2} & \textbf{57.2} & \textbf{75.7} & \textbf{89.9} & \textbf{101.6} & \textbf{109.3} \\
\bottomrule
\end{tabular}
\end{table}

\subsection{Model Complexity Analysis}
In this section, we analyze the time and memory complexity of LuKAN by considering its main architectural components: spatial projections, DWT and its IDWT, and the Temporal Dependency Learner based on KAN with Lucas polynomial activations.

\smallskip\noindent\noindent\textit{Time Complexity.}\quad Each spatial projection involves a matrix multiplication of complexity $\mathcal{O}(DJL)$, where $J$ is the number of joints, $D$ is the embedding dimension, and $L$ is the length of the input sequence. DWT and its inverse are applied along the temporal dimension. As these are linear-time operations per sequence and per feature, their total complexity is $\mathcal{O}(JL)$. The core component of LuKAN is a $B$-layer KAN with Lucas polynomial activations, where $B$ is the total number of blocks. Its time complexity is $\mathcal{O}(BDRL^2)$, where $R$ is the degree of the Lucas polynomial. Hence, the time complexity of LuKAN is $\mathcal{O}(DJL + BDRL^2)$.

\smallskip\noindent\noindent\textit{Memory Complexity.}\quad In terms of memory complexity, the model maintains a lightweight parameter count. Each spatial projection require $\mathcal{O}(JD)$ parameters, while the $B$-layer KAN contributes $\mathcal{O}(BRL^2)$ parameters, giving a total parameter complexity of $\mathcal{O}(JD + BRL^2)$. During runtime, memory is also allocated for storing intermediate activations and for evaluating the polynomial basis, yielding a total runtime memory complexity of $\mathcal{O}(DL + JL)$. Overall, LuKAN achieves a compelling balance between expressive power and computational efficiency.

\section{Conclusion}
In this work, we proposed LuKAN, an effective model for predicting 3D human motion, inspired by Kolmogorov-Arnold networks. Our model captures both localized temporal dependencies and complex motion dynamics effectively. The model's spatial projections ensure that LuKAN maintains structural consistency while remaining computationally efficient. Through extensive experiments on three benchmark datasets, we demonstrated that our model achieves competitive or superior prediction performance compared to state-of-the-art methods, with significantly fewer parameters and lower computational cost. Notably, LuKAN strikes a good balance between prediction accuracy, efficiency, and model simplicity. For future work, we will explore extending LuKAN to handle multi-person scenarios, and further optimizing its architecture for broader applicability.

\bibliography{references}
\clearpage
\appendix
\renewcommand{\thesection}{\Alph{section}}
\renewcommand{\thesubsection}{\Alph{section}.\arabic{subsection}}

\setcounter{page}{1}
\section*{\LARGE{--- Supplementary Material ---\\
LuKAN: A Kolmogorov-Arnold Network Framework for 3D Human Motion Prediction}}

\section{Approach Overview}
The proposed model architecture begins with applying DWT to the input 3D motion sequence. Unlike the discrete cosine transform, which creates a representation of the joint trajectory using cosine waves that oscillate indefinitely, wavelet functions are compact and designed such that their oscillations diminish over time, thereby allowing not only efficient access of localized information about the trajectory, but also capturing rapidly changing features in a trajectory, which DCT cannot address as efficiently. A spatial projection follows, explicitly modeling the spatial structure of the human body by analyzing inter-joint relationships. The core of the architecture is the Temporal Dependency Learner, repeated $B$ times, which consists of a KAN layer parameterized by Lucas polynomials, a layer normalization, and a residual skip connection to effectively capture both local and global temporal patterns. To ensure accurate spatial representation, a second spatial projection is applied after the temporal learner. Then, the data is transformed back to the time domain using the inverse discrete wavelet transform, reconstructing a sequence of predicted 3D human poses.

\section{Dataset Details}
\noindent\textbf{Human3.6M.} is among the most extensive and widely adopted datasets for 3D human motion analysis, comprising over 3.6 million frames of human activities recorded in a controlled indoor environment. It features 7 actors performing 15 everyday activities, including Walking, Sitting, Eating, Greeting, Discussing, and Taking Photos. The dataset is preprocessed to represent poses as 3D coordinates with 22 joints per frame. For evaluation, Subject 11 (S11) is reserved for validation, Subject 5 (S5) is used for testing, and the remaining 5 subjects are employed for training.

\medskip\noindent\textbf{AMASS} is a comprehensive motion database, consolidating multiple optical marker-based motion capture datasets into a standardized framework via the Skinned Multi-Person Linear (SMPL) model. This standardization ensures uniform parameterization across diverse motion recordings. Comprising 51K frames of human activities in indoor and outdoor environments, each pose is represented by 23 joints. For evaluation, the AMASS-BMLrub dataset is used for testing.

\medskip\noindent\textbf{3DPW} includes a diverse range of challenging human activities, spanning both controlled indoor settings and natural outdoor environments. To assess the robustness and generalization of our model trained on AMASS, we evaluate its performance on the 3DPW test set, where each pose is represented by 18 body joints.

\section{Performance and Model Size Comparison}
\begin{figure}[!htb]
\centering\includegraphics[width=3.25in]{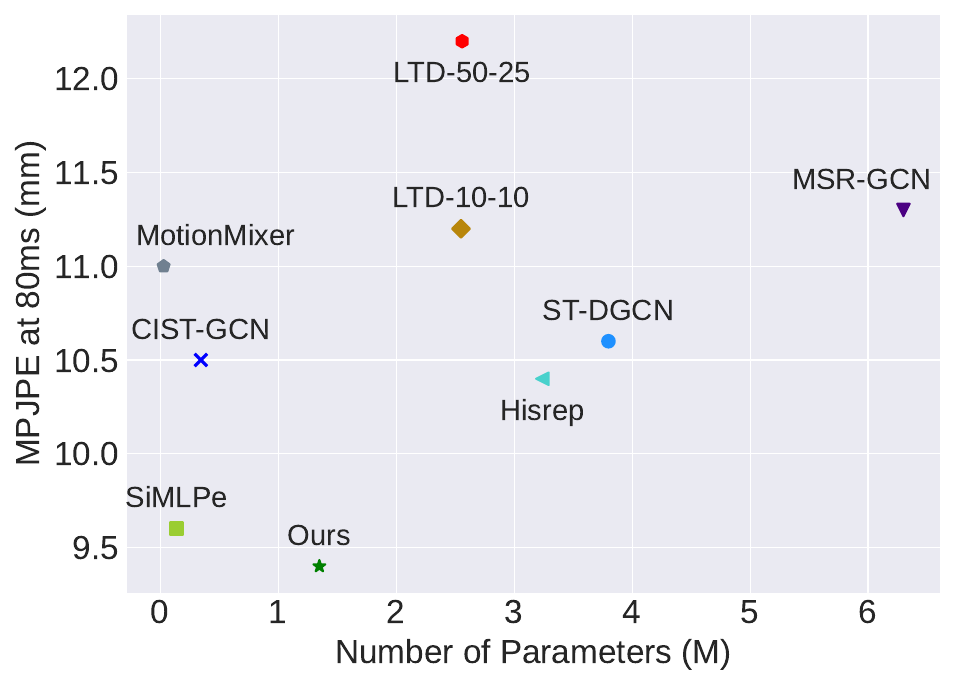}
\caption{\textbf{Comparison of performance and model complexity.} Our model is benchmarked against state-of-the-art methods, including  LTD~\cite{mao2019learning}, Hisrep~\cite{mao2020history}, MSR-GCN~\cite{dang2021msrgcn}, ST-DGCN~\cite{ma2022progressively}, CIST-GCN~\cite{Medina2024CISTGCN}, MotionMixer~\cite{Arij2022MotionMixer}, and SiMLPe~\cite{guo2023mlp}. Performance is assessed using the Mean Per Joint Position Error (MPJPE), where lower values indicate superior prediction performance. All evaluations are performed on the Human3.6M dataset.}
\label{Fig:ModelSize}
\end{figure}

A significant advantage of our LuKAN model lies in its simplicity, which distinguishes it from many existing methods for 3D human motion prediction. LuKAN's simple design eliminates unnecessary complexity, focusing instead on robust modeling of temporal and spatial dependencies. Despite its straightforward architecture, LuKAN achieves state-of-the-art performance, demonstrating reduced prediction errors compared to strong baseline methods. As illustrated in Figure~\ref{Fig:ModelSize}, LuKAN achieves competitive performance by reducing Mean Per Joint Position Error (MPJPE), outperforming strong baseline methods such as LTD~\cite{mao2019learning}, Hisrep~\cite{mao2020history}, MSR-GCN~\cite{dang2021msrgcn}, ST-DGCN~\cite{ma2022progressively}, CIST-GCN~\cite{Medina2024CISTGCN}, MotionMixer~\cite{Arij2022MotionMixer}, and SiMLPe~\cite{guo2023mlp}. This comparison, conducted on the widely used Human3.6M dataset, highlights LuKAN's ability to deliver accurate motion predictions while maintaining a compact model size. The simplicity of LuKAN translates directly into computational efficiency, making it an ideal solution for applications requiring lightweight models without compromising prediction performance.

\section{Experimental Results}
\noindent\textbf{Action-Wise Performance.}\quad The results in Table~\ref{tab:short_term_prediction} demonstrate that our LuKAN model consistently outperforms or matches the performance of the best-performing baselines, MotionMixer and SiMLPe, across most actions and prediction horizons in terms of MPJPE. For instance, in the Eating action at 400ms, LuKAN achieves an MPJPE of 35.5mm compared to 36.1mm for SiMLPe, resulting in a relative error reduction of 1.7\%. Similarly, for the Greeting action at 400ms, LuKAN reduces the MPJPE to 76.2mm compared to SiMLPe's 77.3mm, yielding a relative error reduction of approximately 1.42\%, highlighting LuKAN's ability to generate accurate predictions in challenging scenarios.  For the Waiting action at 400ms, LuKAN achieves an MPJPE of 53.3mm, which is comparable to SiMLPe's best result of 53.2mm, showcasing LuKAN's ability to deliver competitive performance even in actions where SiMLPe excels. LuKAN's performance is particularly notable in the Sitting Down action at 320ms, where it achieves an MPJPE of 58.7mm, outperforming MotionMixer's MPJPE of 61.4mm with a relative error reduction of 4.4\%. This significant improvement underscores LuKAN's capability to handle challenging motions with complex joint interactions.

When considering the average errors reported in the bottom-right corner of the table, the results indicate a consistent improvement in accuracy across all prediction horizons. The relative error reductions for the average MPJPE of LuKAN compared to SiMLPe at 320ms and 400ms are approximately 0.22\% and 0.17\%, respectively. Overall, LuKAN shows consistent superiority or parity with the best performing baselines across most actions and prediction horizons, reinforcing its effectiveness in predicting 3D human motion.

\begin{table*}[!htb]
\caption{\textbf{Action-wise performance comparison of our model and baseline methods on Human3.6M} for different prediction horizons ranging from 80ms to 400ms. MPJPE errors are in mm. The average errors are reported in the bottom-right corner of the table.}
\label{tab:short_term_prediction}
\smallskip
\centering
\small
\setlength\tabcolsep{1pt} % Reduce column padding
\begin{tabular}{lcccc|cccc|cccc|cccc}
\toprule
 & \multicolumn{4}{c}{Walking} & \multicolumn{4}{c}{Eating} & \multicolumn{4}{c}{Smoking} & \multicolumn{4}{c}{Discussion} \\
%\cline{2-17}
\cmidrule(lr){2-5}  \cmidrule(lr){6-9}  \cmidrule(lr){10-13}  \cmidrule(lr){14-17}
 & 80 & 160 & 320 & 400 & 80 & 160 & 320 & 400 & 80 & 160 & 320 & 400 & 80 & 160 & 320 & 400\\
\hline
%Res. sup.\cite{ref:ressup}	& 23.2	& 40.9	& 61	& 66.1 & 16.8	& 31.5	& 53.5	& 61.7 & 18.9	& 34.7	& 57.5	& 65.4 & 25.7	& 47.8	& 80	& 91.3\\
ConvSeq2Seq~\cite{li2018convolutional} & 17.7 & 33.5 & 56.3	& 63.6 & 11.0 & 22.4 & 40.7 & 48.4 & 11.6	& 22.8 & 41.3 & 48.9 & 17.1	& 34.5 & 64.8 & 77.6\\
DMGNN~\cite{Maosen2020DMGNN} & 17.3 & 30.7 & 54.6 & 65.2 & 11.0 & 21.4 & 36.2 & 43.9 & 9.0 & 17.6 & 32.1 & 40.3 & 17.3 & 34.8 & 61.0 & 69.8 \\
MSR-GCN~\cite{dang2021msrgcn} & 12.2 & 22.7 & 38.6 & 45.2 & 8.4 & 17.1 & 33.0 & 40.4 & 8.0 & 16.3 & 31.3 & 38.2 & 12.0 & 26.8 & 57.1 & 69.7 \\
LTD-10-10~\cite{mao2019learning} & 11.1	& 21.4	& 37.3	& 42.9 & 7.0	& 14.8	& 29.8	& 37.3 & 7.5 & 15.5	& 30.7 & 37.5 & 10.8 & 24 & 52.7 & 65.8\\
% PGBIG [24] & 10.2 & 19.8 & 34.5 & 40.3 & 7.0 & 15.1 & 30.6 & 38.1 & 6.6 & 14.1 & 28.2 & 34.7 & 10.0 & 23.8 & 53.6 & 66.7 \\
% SPGSN [21] & 10.1 & 19.4 & 34.8 & 41.5 & 7.1 & 14.9 & 30.5 & 37.9 & 6.7 & 13.8 & 28.0 & 34.6 & 10.4 & 23.8 & 53.6 & 67.1 \\

Hisrep~\cite{mao2020history} & 10.0 &	\underline{19.5} & \textbf{34.2} & \underline{39.8} & 6.4 &	\underline{14.0} & 28.7 & 36.2 & 7.0 & 14.9 & 29.9 & 36.4 & 10.2 & 23.4 & 52.1 & 65.4\\
MotionMixer~\cite{Arij2022MotionMixer}	& 10.8	& 22.4	& 36.5	& 42.4 & 7.7 & \underline{14.0} & \textbf{27.3} & \underline{36.1} & 7.1 & \textbf{14.0} & \textbf{29.1} & 36.8 & 10.2 & \underline{22.5}	& \underline{51.0} & \underline{64.1}\\
% \textbf{PTSMXR-R1L80} & \underline{10.2}	& \textbf{19.5}	& \textbf{33.7}	& \textbf{39} & \underline{6.3}	& \underline{13.9}	& 28.4	& \underline{35.6} & 7.1	 & 14.6	& \underline{29.4}	 & 36.5 & -	
% & -	& -	& -\\
SiMLPe~\cite{guo2023mlp} & \underline{9.9}	& -	& -	& \textbf{39.6} & \underline{5.9}	& -	& -	& \underline{36.1} & \underline{6.5}	& -	& - &	 \textbf{36.3}& \underline{9.4}	 & - & -	 & 64.3\\
\hline
% our model (DCT) & \textbf{9.8}	& \textbf{19.3}	& \underline{34.3}	& 40.1 & \textbf{5.8}	& \textbf{13.2}	& \underline{27.7}	& \textbf{35.2} & \textbf{6.5}	& \underline{14.2}	 & \textbf{29} &	\textbf{36} & \textbf{9}	& \textbf{21.8}	& \textbf{50.2} &	\textbf{63.3}\\
Ours & \textbf{9.8}	& \textbf{19.2}	& \underline{34.3}	& 40 & \textbf{5.8}	& \textbf{13.3}	& \underline{28.1}	& \textbf{35.5} & \textbf{6.4}	& \underline{14.1}	& \underline{29.3} &	 \underline{36.4} & \textbf{9.1} &	\textbf{22.2} &	\textbf{50.4} & \textbf{63.7}\\

\toprule
 & \multicolumn{4}{c}{Directions} & \multicolumn{4}{c}{Greeting} & \multicolumn{4}{c}{Phoning} & \multicolumn{4}{c}{Posing} \\
\cmidrule(lr){2-5}  \cmidrule(lr){6-9}  \cmidrule(lr){10-13}  \cmidrule(lr){14-17}
 & 80 & 160 & 320 & 400 & 80 & 160 & 320 & 400 & 80 & 160 & 320 & 400 & 80 & 160 & 320 & 400\\
\hline
%Res. sup.\cite{ref:ressup}	& 21.6	& 41.3	& 72.1	& 84.1 & 31.2 & 58.4	& 96.3	& 108.8  & 21.1	& 38.9	& 66	& 76.4 & 29.3	& 56.1 &	98.3	 & 114.3\\
ConvSeq2Seq~\cite{li2018convolutional} & 13.5 & 29.0 & 57.6 & 69.7 & 22.0 & 45.0 & 82.0 & 96.0 & 13.5 & 26.6 & 49.9 & 59.9 & 16.9 & 36.7 & 75.7 & 92.9\\
DMGNN~\cite{Maosen2020DMGNN} & 13.1 & 24.6 & 64.7 & 81.9 & 23.3 & 50.3 & 107.3 & 132.1 & 12.5 & 25.8 & 48.1 & 58.3 & 15.3 & 29.3 & 71.5 & 96.7 \\
MSR-GCN~\cite{dang2021msrgcn} & 8.6 & 19.7 & \underline{43.3} & \underline{53.8} & 16.5 & 37.0 & 77.3 & 93.4 & 10.1 & 20.7 & 41.5 & 51.3 & 12.8 & 29.4 & 67.0 & 85.0 \\
% PGBIG [24] & 7.2 & 17.6 & 40.9 & 51.5 & 15.2 & 34.1 & 71.6 & 87.1 & 8.3 & 18.3 & 38.7 & 48.4 & 10.7 & 25.7 & 60.0 & 76.6 \\
% SPGSN [21] & \underline{7.4} & \underline{17.1} & 39.8 & 50.3 & 14.6 & 32.6 & 70.6 & 86.4 & 8.7 & 18.3 & 38.7 & 48.5 & 10.7 & 25.3 & 59.9 & 76.5 \\

LTD-10-10~\cite{mao2019learning} & 8.0 & 18.8 & 43.7 & 54.9 & 14.8 & 31.4 & 65.3 & 79.7 & 9.3 & 19.1 & 39.8 & 49.7 & 10.9 & 25.1 & 59.1 & 75.9\\
Hisrep~\cite{mao2020history} & \underline{7.4} & 18.4 & 44.5 & 56.5 & 13.7 & \underline{30.1} & 63.8 & 78.1 & 8.6 & \underline{18.3} & 39.0 & 49.2 & 10.2 & 24.2 & \underline{58.5}	 & 75.8\\
MotionMixer~\cite{Arij2022MotionMixer} & 8.3 & \underline{18.1} & 43.8 & \textbf{53.4} & 12.8 & 33.4 & \underline{62.3}	& 82.2 & 10.0	& 20.1 & \textbf{37.4} & 51.1 & 11.7	& \underline{23.3} & 62.4 & 79.5\\
SiMLPe~\cite{guo2023mlp} & \textbf{6.5}	& -	& -	& 55.8 & \underline{12.4} & - & - & \underline{77.3} & \underline{8.1} & - & - & \underline{48.6} & \underline{8.8} & - & - & \textbf{73.8}\\
\hline
% \textbf{PTSMXR-R1L80} & & & & & \underline{13.1} &	\textbf{28.5}	& \textbf{61.3} & \textbf{75.2} & 8.7	& \textbf{18.1}	& \underline{38.2}	& \textbf{48.2} & \underline{9.8}	 & \underline{23.3}	& \textbf{55.9}	& \textbf{72.2}\\
% our model (DCT) & \textbf{6.4}	& \textbf{17}	& \textbf{43}	& 54.7 & \textbf{12.1}	& \textbf{28.3}	& \textbf{62}	& \textbf{76.3} & \textbf{7.8} &	 \textbf{17.3}	& \underline{37.6} &	\textbf{47.2} & \textbf{8.5}	& \textbf{22.3}	& \textbf{56.1}	& \textbf{72.4}\\
Ours & \textbf{6.5} & \textbf{17.3} & \textbf{43.1} & 54.9 & \textbf{12.1} & \textbf{28.6} &	\textbf{62.2} &	\textbf{76.2} & \textbf{7.9} &	 \textbf{17.5} & \underline{38.3} & \textbf{48.1} & \textbf{8.5}	& \textbf{22.5}	& \textbf{57.3}	& \underline{74.1}\\

\toprule
 & \multicolumn{4}{c}{Purchases} & \multicolumn{4}{c}{Sitting} & \multicolumn{4}{c}{Sitting Down} & \multicolumn{4}{c}{Taking Photo} \\
\cmidrule(lr){2-5}  \cmidrule(lr){6-9}  \cmidrule(lr){10-13}  \cmidrule(lr){14-17}
 & 80 & 160 & 320 & 400 & 80 & 160 & 320 & 400 & 80 & 160 & 320 & 400 & 80 & 160 & 320 & 400\\
\hline
%Res. sup.\cite{ref:ressup}	& 28.7	& 52.4	& 86.9	& 100.7 & 23.8	& 44.7	& 78	& 91.2 & 31.7	& 58.3	& 96.7 & 112 & 21.9	& 41.4	& 74	& 87.6\\
ConvSeq2Seq~\cite{li2018convolutional} & 20.3 & 41.8 & 76.5	& 89.9 & 13.5 & 27.0 & 52.0	& 63.1 & 20.7 & 40.6 & 70.4	& 82.7 & 12.7 & 26.0 & 52.1 & 63.6\\
DMGNN~\cite{Maosen2020DMGNN} & 21.4 & 38.7 & 75.7 & 92.7 & 11.9 & 25.1 & 44.6 & 50.2 & 15.0 & 32.9 & 77.1 & 93.0 & 13.6 & 29.0 & 46.0 & 58.8 \\
MSR-GCN~\cite{dang2021msrgcn} & 14.8 & 32.4 & 66.1 & 79.6 & 10.5 & 22.0 & 46.3 & 57.8 & 16.1 & 31.6 & 62.5 & 76.8 & 9.9 & 21.0 & 44.6 & 56.3 \\
% PGBIG [24] & 12.5 & 28.7 & 60.1 & 73.3 & 8.8 & 19.2 & 42.4 & 53.8 & 13.9 & 27.9 & 57.4 & 71.5 & 8.4 & 18.9 & 42.0 & 53.3 \\
% SPGSN [21] & 12.8 & 28.6 & 61.0 & 74.4 & 9.3 & 19.4 & 42.3 & 53.6 & 14.2 & 27.7 & 56.8 & 70.7 & 8.7 & 18.9 & 41.5 & 52.7 \\
LTD-10-10~\cite{mao2019learning} & 13.9	& 30.3 & 62.2 & 75.9 & 9.8 & 20.5 & 44.2 & 55.9 & 15.6 & 31.4 & \underline{59.1} & \underline{71.7} & 8.9 & 18.9  & 41.0 & 51.7\\
Hisrep~\cite{mao2020history} & \underline{13.0} & \underline{29.2} & \underline{60.4} & 73.9 & \underline{9.3} & \underline{20.1} & 44.3 & 56.0 & 14.9 & \underline{30.7} & \underline{59.1} & 72.0 & 8.3 & \underline{18.4}	& \underline{40.7} & 51.5\\
MotionMixer~\cite{Arij2022MotionMixer} & 14.6 & 31.3 & 62.8	& 76.1 & 10.0	& 20.9 & \textbf{43.7} & \textbf{54.5} & \textbf{12.0} & 31.4 & 61.4	& 74.5 & 9.0 & 18.9	& 41.0 & 51.6\\
SiMLPe~\cite{guo2023mlp} & \textbf{11.7} & - & - & \textbf{72.4} & \textbf{8.6} & - & - & \underline{55.2} & \underline{13.0} & - & -	& \textbf{70.8} & \underline{7.8} & - & - & \textbf{50.8} \\
\hline
% \textbf{PTSMXR-R1L80} & \underline{12.6}	& \underline{28.8}	& \textbf{60.1}	& \textbf{73.7} & \underline{9.3}	& \underline{19.9}	& \underline{43.7}	 & 55.6 & 14.8	& 30.8	 & 59.3	& 72.3 & \textbf{8.2}	& \textbf{18.1}	& \textbf{39.9}	& \textbf{50.6}\\
% our model (DCT) & \textbf{11.4}	& \textbf{27.5}	& \textbf{58.9}	& \underline{72.5} & \textbf{8.5}	& \textbf{19}	& \textbf{42.8}	& \textbf{54.4} & 13.5 & 	 \textbf{29.4} &	 \textbf{58}	& \textbf{70.5} & \textbf{7.6}	& \textbf{17.6}	& \textbf{40.1}	& \underline{51} \\
Ours & \textbf{11.7} & \textbf{27.8} & \textbf{59.2} & \underline{72.9} & \textbf{8.6}	& \textbf{19.4}	& \underline{43.9} & 55.7 & 13.6 &	 \textbf{29.5} & \textbf{58.7} &	 71.9 & \textbf{7.7} & \textbf{ 17.7} &	\textbf{40.2} & \underline{51.1} \\

\toprule
& \multicolumn{4}{c}{Waiting} & \multicolumn{4}{c}{Walking Dog} & \multicolumn{4}{c}{Walking Together} & \multicolumn{4}{c}{Average} \\
\cmidrule(lr){2-5}  \cmidrule(lr){6-9}  \cmidrule(lr){10-13}  \cmidrule(lr){14-17}
 & 80 & 160 & 320 & 400 & 80 & 160 & 320 & 400 & 80 & 160 & 320 & 400 & 80 & 160 & 320 & 400\\
\hline
%Res. sup.\cite{ref:ressup}	& 23.8	& 44.2	& 75.8	& 87.7 & 36.4	& 64.8	& 99.1	& 110.6 & 20.4	& 37.1	& 59.4	& 67.3 & 25.0 & 46.2 & 77.0 & 88.3\\
ConvSeq2Seq~\cite{li2018convolutional} & 14.6 & 29.7 & 58.1 & 69.7 & 27.7 & 53.6 & 90.7	& 103.3 & 15.3 & 30.4 & 53.1	& 61.2 & 16.6 & 33.3 & 61.4 & 72.7\\
DMGNN~\cite{Maosen2020DMGNN} & 12.2 & 24.2 & 59.6 & 77.5 & 47.1 & 93.3 & 160.1 & 171.2 & 14.3 & 26.7 & 50.1 & 63.2 & 17.0 & 33.6 & 65.9 & 79.7 \\
MSR-GCN~\cite{dang2021msrgcn}  & 10.7 & 23.1 & 48.3 & 59.2 & 20.7 & 42.9 & 80.4 & 93.3 & 10.6 & 20.9 & 37.4 & 43.9 & 12.1 & 25.6 & 51.6 & 62.9 \\
LTD-10-10~\cite{mao2019learning} & 9.2 & 19.5 & \underline{43.3} & 54.4 & 20.9 & 40.7 & 73.6 & 86.6 & 9.6 & 19.4 & 36.5 & 44 & 11.2 & 23.4 & 47.9 & 58.9\\
Hisrep~\cite{mao2020history} & 8.7 & \underline{19.2} & 43.4 & 54.9 & 20.1 & \underline{40.3} & \underline{73.3} & 86.3 & 8.9 & \underline{18.4} & \underline{35.1} & 41.9 & 10.4 & 22.6 & 47.1 & 58.3\\
MotionMixer~\cite{Arij2022MotionMixer} & 10.2 & 21.1 & 45.2	& 56.4 & 20.5 & 42.8 & 75.6	& 87.8 & 10.5 & 20.6 & 38.7	& 43.5 & 11.0 & 23.6 & 47.8 & 59.3\\
SiMLPe~\cite{guo2023mlp} & \underline{7.8} & -	& -	& \textbf{53.2} & \underline{18.2} & - & - & \underline{83.6} & \underline{8.4} & -	& -	& \underline{41.2} & \underline{9.6} & \underline{21.7} & \underline{46.3} & \underline{57.3} \\
\hline
% \textbf{PTSMXR-R1L80} & \underline{8.4}	& \textbf{18.3}	& \underline{41.6}	& \underline{52.9} & \textbf{18.9}	& \textbf{38.8}	& \textbf{70.9}	& \textbf{83.9} & \underline{8.9}	 & \underline{17.9}	& \underline{33.7}	& \underline{40.8} & \underline{10.4}	& \underline{22.4}	& \underline{46.6}	 & \underline{57.4}\\
% our model (DCT) & \textbf{7.5}	& \textbf{17.7}	& \textbf{41.1}	& \textbf{52.5} & \textbf{18.1} & \textbf{38.6} & \textbf{72.1} & \underline{85.3} & \textbf{8.2}	& \textbf{17.5}	 & \textbf{33.9}	& \textbf{40.5} & \textbf{9.4}	& \textbf{21.4}	& \textbf{45.8}	& \textbf{56.8} \\
Ours & \textbf{7.6}	& \textbf{17.9}	& \textbf{42} & \underline{53.3} & \textbf{18.1} & \textbf{38.1} &	\textbf{71} & \textbf{83.3} & \textbf{8.2} & \textbf{17.6} & \textbf{34.4} & \textbf{41.1} & \textbf{9.4} & \textbf{21.5} & \textbf{46.2} & \textbf{57.2} \\

% PGBIG [24] & 8.9 & 20.1 & 43.6 & 54.3 & 18.8 & 39.3 & 73.7 & 86.4 & 8.7 & 18.6 & 34.4 & 41.0 & 10.3 & 22.7 & 47.4 & 58.5 \\
% SPGSN [21] & 9.2 & 19.8 & 43.1 & 54.1 & 18.2 & 37.3 & 71.3 & 84.2 & 8.9 & 18.2 & 33.8 & 40.9 & 10.4 & 22.3 & 47.1 & 58.3 \\
% \textbf{Ours} & \textbf{8.2} & \textbf{18.4} & \textbf{41.3} & \textbf{52.1} & \textbf{14.5} & \textbf{32.7} & \textbf{63.8} & \textbf{76.0} & \textbf{7.4} & \textbf{15.2} & \textbf{30.0} & \textbf{36.4} & \textbf{9.3} & \textbf{19.7} & \textbf{41.0} & \textbf{51.1} \\
\bottomrule
\end{tabular}
\end{table*}

\section{Ablation Study}
\noindent\textbf{Effect of Embedding Dimension.}\quad Table~\ref{tab:spatial_dim} reports the effect of varying the embedding dimension $D$ of the spatial projection on model prediction in terms of MPJPE across different time steps using the Human3.6M dataset. In this dataset, each 3D human pose comprises 22 joints, represented by 3 coordinates $(x, y, z)$, resulting in an input spatial dimension of $D=66$. This input is processed by the spatial projection layer, which maps the poses into different embedding dimensions to optimize the representation for the 3D human motion prediction task. By varying the embedding dimension $D$, the spatial projection layer refines the inter-joint dependencies, ensuring the spatial relationships are effectively captured for accurate motion prediction. The best performance is observed at $D=200$, which achieves an MPJPE of 109.3mm at 1000ms, outperforming $D=66$ with 110.3mm. The consistent improvement across all time steps demonstrates that setting $D=200$ provides a balanced representation, enhancing the model's ability to capture spatial dependencies effectively, while larger or smaller dimensions introduce insufficient expressiveness.

\begin{table}[!htb]
\caption{Ablation study on the embedding dimension $D$ of the spatial projection for various prediction horizons. The best performance is achieved with $D=200$.}
\label{tab:spatial_dim}
\smallskip
\centering
\small
\setlength\tabcolsep{2.4pt} % Reduce column padding
\begin{tabular}{lcccccccc}
\toprule
\multirow{2}{*}{Embed Dim. ($D$)} & \multicolumn{8}{c}{MPJPE (mm)$\downarrow$} \\
\cmidrule(lr){2-9}
& 80 & 160 & 320 & 400 & 560 & 720 & 880 & 1000 \\
\midrule
66 & 9.8 & 22.2 & 47.0 & 58.1 & 76.7 & 90.7 & 102.5 & 110.3\\
100 & 9.6 & 21.8 & 46.5 & 57.4 & 76.1 & 90.5 & 102.4 & 110.1\\
150 & 9.5 & 21.7 & 46.5 & 57.5 & 76.0 & 90.3 & 101.9 & 109.6 \\
200 & \textbf{9.4} & \textbf{21.5} & \textbf{46.2} & \textbf{57.2} & \textbf{75.7} & \textbf{89.9} & \textbf{101.6} & \textbf{109.3} \\
220 & 9.6 & 21.7 & 46.3 & 57.2 & 76.0 & 90.6 & 102.6 & 110.4\\
\bottomrule
\end{tabular}
\end{table}

\section{Discussion}
This work presents LuKAN, an effective and interpretable model for 3D human motion prediction that strikes a balance between simplicity and accuracy while effectively capturing spatio-temporal dependencies.  Despite its promising contributions, several limitations and avenues for future research deserve consideration.

\medskip\noindent\textbf{Limitations.}\quad While LuKAN achieves competitive performance and computational efficiency, its robustness on large-scale and highly diverse motion datasets containing noisy or incomplete data requires further investigation. In addition, although the model is lightweight, further optimization could be explored enhance its suitability for real-time applications, especially in resource-constrained environments.

\medskip\noindent\textbf{Future Work.}\quad Extending LuKAN to multi-person scenarios or incorporating contextual factors such as environmental constraints could improve its performance in complex, real-world settings. Future efforts could also focus on scaling the model to handle longer motion sequences or higher-dimensional data, expanding its applicability. Exploring alternative polynomial bases may uncover configurations that further enhance accuracy and generalization. Furthermore, integrating LuKAN with unsupervised or semi-supervised learning paradigms could reduce its dependence on large-scale annotated datasets, broadening its accessibility for diverse applications.

\medskip\noindent\textbf{Social Impact.}\quad LuKAN offers substantial potential for positive societal impact through accurate, efficient, and interpretable 3D human motion prediction. In healthcare, LuKAN could improve physical rehabilitation systems by predicting patient movements, leading to better therapy outcomes. In autonomous systems, LuKAN could enhance human-robot interaction by enabling robots to anticipate human actions, improving safety and usability. In the entertainment and animation industries, its lightweight design could support real-time character animation, lowering production costs and increasing accessibility for smaller studios. In adddition, the interpretability of LuKAN fosters trust and transparency, ensuring ethical use in sensitive domains such as surveillance and monitoring.  
\end{document}